\documentclass{article}

\usepackage{arxiv}

\usepackage[utf8]{inputenc}
\usepackage[T1]{fontenc} 
\usepackage{hyperref}
\usepackage{url}

\usepackage{microtype}
\usepackage{graphicx}
\usepackage{wrapfig}
\usepackage{subfigure}
\usepackage{booktabs}

\usepackage{amsfonts, amsmath}
\usepackage{amssymb}
\usepackage{mathtools}
\usepackage{amsthm}

\usepackage{nicefrac}
\usepackage[round]{natbib}

\usepackage{xcolor}

\usepackage{doi}
\usepackage{calrsfs}
\usepackage{bm}
\DeclareMathAlphabet{\pazocal}{OMS}{zplm}{m}{n}
\usepackage{array}
\usepackage{multirow}
\usepackage{booktabs}

\makeatletter
\g@addto@macro{\endtabular}{\rowfont{}}
\makeatother
\newcommand{\rowfonttype}{}
\newcommand{\rowfont}[1]{
   \gdef\rowfonttype{#1}#1%
}
\newcolumntype{L}[1]{>{\raggedright\let\newline\\\arraybackslash\rowfonttype}m{#1}}
\newcolumntype{C}[1]{>{\centering\let\newline\\\arraybackslash\rowfonttype}m{#1}}
\newcolumntype{R}[1]{>{\raggedleft\let\newline\\\arraybackslash\rowfonttype}m{#1}}
\usepackage{pgfplots}
\usetikzlibrary{arrows.meta}
\usepgfplotslibrary{groupplots}
\usepgfplotslibrary{dateplot}
\usepackage[labelfont=bf]{caption}
\usepackage{float}

\definecolor{blue(pigment)}{rgb}{0.2, 0.2, 0.6}
\definecolor{darkGreen}{rgb}{0.0, 0.5, 0.0}
\usepackage{mfirstuc}
\MFUnocap{for}
\MFUnocap{or}
\MFUnocap{etc}
\usepackage{stfloats}

\pgfplotsset{every axis/.append style={
                    label style={font=\normalsize},
                    title style={font=\normalsize},
                    tick label style={font=\normalsize}  
                    }}

\title{\textbf{\capitalisewords{Self-attention presents low-dimensional knowledge graph embeddings for link prediction}}}

\author{{Peyman Baghershahi}\\
	University of Tehran\\
	Tehran, Iran \\
	\texttt{p.baghershahi@ut.ac.ir} \\
	\And
	{Reshad Hosseini} \\
	University of Tehran\\
	Tehran, Iran \\
	\texttt{reshad.hosseini@ut.ac.ir} \\
	\And
	{Hadi Moradi} \\
	University of Tehran\\
	Tehran, Iran \\
	\texttt{moradih@ut.ac.ir} \\
}

\renewcommand{\shorttitle}{}

\hypersetup{
pdftitle={A template for the arxiv style},
pdfsubject={q-bio.NC, q-bio.QM},
pdfauthor={David S.~Hippocampus, Elias D.~Striatum},
pdfkeywords={First keyword, Second keyword, More},
colorlinks=true,
citecolor={blue(pigment)},
citebordercolor={blue(pigment)},
urlcolor={blue(pigment)},
}

\begin{document}
\maketitle

\begin{abstract}
A few models have tried to tackle the link prediction problem, also known as knowledge graph completion, by embedding knowledge graphs in comparably lower dimensions. However, the state-of-the-art results are attained at the cost of considerably increasing the dimensionality of embeddings which causes scalability issues in the case of huge knowledge bases. Transformers have been successfully used recently as powerful encoders for knowledge graphs, but available models still have scalability issues. To address this limitation, we introduce a Transformer-based model to gain expressive low-dimensional embeddings. We utilize a large number of self-attention heads as the key to applying query-dependent projections to capture mutual information between entities and relations. Empirical results on WN18RR and FB15k-237 as standard link prediction benchmarks demonstrate that our model has favorably comparable performance with the current state-of-the-art models. Notably, we yield our promising results with a significant reduction of 66.9\% in the dimensionality of embeddings compared to the five best recent state-of-the-art competitors on average.\footnote{Code available at \url{https://github.com/Peyman-Bi/SAttLE}}
\end{abstract}

\section{Introduction}
Knowledge Graphs (KGs) are widely used in various applications such as question answering, information retrieval, recommendation systems, and web search. However, KGs are commonly incomplete with lots of missed relations. Link prediction, also called KG completion, is a kind of automated reasoning to predict missing links between entities of KGs. Existing link prediction methods on KGs lie in two principal directions: embedding-based methods and neural network methods. 

Embedding-based methods are faster and have lower nonembedding free parameters. Two main categories of them are translational and semantic matching methods. Translational methods \citep{NIPS2013_1cecc7a7, wang2014knowledge, sun2018rotate, NEURIPS2019_f8b932c7} learn linear translations from source to target entities considering their relations. Semantic matching (or tensor decomposition) methods match latent semantics of entities with respect to relations \citep{DBLP:conf/icml/NickelTK11, DBLP:journals/corr/YangYHGD14a, trouillon2016complex, lacroix2018canonical, NEURIPS2019_d961e9f2}. However, embedding-based methods have scalability issues facing huge KGs due to high dimensional embeddings required for encoding their whole information. This also can lead to overfitting and complexity overload. Besides, they mainly utilize additive or multiplicative operations which result in lack of expressiveness.

In contrast, neural network approaches \citep{socher2013reasoning, schlichtkrull2018modeling, dettmers2018convolutional, bansal2019a2n, schlichtkrull2018modeling} enrich representations by storing some learned knowledge in nonembedding free parameters of the model to be shared through the whole KG. However, most of these methods use shallow or inefficient structures that cannot exploit mutual interactions and dependencies between entities and relations to increase expressiveness of representation. Therefore, existing models almost achieve their best results using high dimensional embeddings.

Recently, Transformers \citep{NIPS2017_3f5ee243} have been successfully used as general encoders for knowledge graph completion with promising results \citep{wang2019:coke, chen-etal-2021-hitter}. This is because of the ability of self-attention to effectively capture the mutual information and dependencies within a sequence \citep{shen2019mutual}. Indeed, entities and relations representations should vary based on the graph context they appear in \citep{wang2019:coke}. Thus, the self-attention matrix applies query-dependent linear projections to a pair of source entity and relation as the input query in order to propagate dependencies and interactions between them and generate context-aware representations rather than just memorizing general static representations.

Although previous Transformer-based models achieved considerably better results compared to previous models with approximately the same high dimensional embeddings, they still have scalability issues. Moreover, they are large models and use multiple encoder blocks which impose too many nonembedding free parameters and increase time and memory complexity considerably. We introduce SAttLE, a Transformer-based model, to tackle the aforementioned limitations. We found the significance of the number of self-attention heads for capturing dependencies between entities and relations. In contrast to previous similar models, we use a single Transformer encoder block, rather than multiple ones, but with numerous heads to notably reduce free parameters. Consequently, we attain competitive results with embeddings in remarkably low dimensions by stacking simple decoders to our powerful general encoder which makes our model extremely scalable.

In summary, our key contributions are as follows:
\begin{itemize}
    \item We introduced SAttLE, a highly expressive Transformer-based knowledge graph encoder to address the scalability issue on enormous KGs.
    \item The importance of self-attention heads is investigated and the model size is reduced by a new structure composed of a single Transformer encoder with numerous heads.
    \item we achieve competitive performance with state-of-the-art models on FB15k-237 and WN18RR, as standard link prediction benchmarks, by a significant reduction of 66.9\% in the dimensionality of embeddings compared to the five best competitors on average.
\end{itemize}

\section{Related Works}

Generally, there are two broad categories in the field of link prediction on KGs: embedding-based methods and neural network methods. Embedding-based methods score triples using computed embeddings for entities and relations; They contain two subcategories: translational and semantic matching approaches.

\subsection{Embedding-based methods}

TransE \citep{NIPS2013_1cecc7a7} can be considered as the earliest translational method which considers a simple translation from source to target parameterized by relation, and TransH \citep{wang2014knowledge} extends these translations to hyperplanes. On the other hand, semantic matching methods try to tie latent semantics of entities with respect to relations. In this category, there are bilinear methods such as RESCALE \citep{DBLP:conf/icml/NickelTK11} and DistMult \citep{DBLP:journals/corr/YangYHGD14a}, both with single embedding for each entity, and similarly SimplE \citep{NEURIPS2018_b2ab0019} based on Canonical Polyadic (CP) decomposition \citep{hitchcock1927expression} except that it dependently embeds entities as source or target with respect to relations and their inverses. Also, TuckER \citep{balazevic-etal-2019-tucker} based on Tucker decomposition \citep{tucker1966some} with a trainable core tensor utilize multi-task learning to share information. Apart from methods using multiplication, HOLE \citep{nickel2016holographic} suggested circular correlation to create compositional representations. ComplEx \citep{trouillon2016complex} first noted the capability of complex space to capture anti-symmetric relations. In the same domain, ComplEx-N3 \citep{lacroix2018canonical} utilized reciprocal learning and a new regularization method. In order to infer relations such as (anti-)symmetry and inversion, RotatE \citep{sun2018rotate} defined relations as rotations from source to target entity on a single complex plane, and QuatE \citep{NEURIPS2019_d961e9f2} generalized these rotations to two planes by quaternions. Besides, some methods such as MuRP \citep{NEURIPS2019_f8b932c7} and ATTH \citep{chami-etal-2020-low} that encode KGs in hyperbolic space due to its efficiency for embedding hierarchical data.

Embedding-based methods are not usually applicable to enormous KGs because they need to increase KGs embeddings dimensionality to enhance their expressiveness. So, they face scalability issues and, in cases, overfitting due to high complexity.

\subsection{Neural network methods}

Contrary to embedding methods, neural network methods obtain expressive representations from pure embeddings using different neural network models. More specifically, NTN \citep{socher2013reasoning} first proposed a simple feed-forward neural tensor network. Later, ConvE \citep{dettmers2018convolutional} got effective results, using a Convolutional Neural Network (CNN) on interactive 2d reshaped embeddings. Recently, inspired by ConvE, InteractE \citep{vashishth2020interacte} increased the number of interactions between relation and entity resulting in significant improvements. Similarly, HypER \citep{balavzevic2019hypernetwork} uses CNNs with relation-specific filter weights for multi-task learning. 

Furthermore, Graph Neural Network (GNN) models such as R-GCN \citep{schlichtkrull2018modeling}, and COMPGCN \citep{Vashishth2020Composition-based}, take advantage of graph structure and entities neighborhood to obtain better representations. Similar to GNN models, A2N \citep{bansal2019a2n} also represents entities based on query-conditioned bilinear attention on their graph neighbors. 

Recently, Transformers \citep{NIPS2017_3f5ee243} have made promising improvements in various problems \citep{devlin-etal-2019-bert, dosovitskiy2021an} as well as KG completion. KG-BERT \citep{DBLP:journals/corr/abs-1909-03193} fine-tunes BERT pre-trained language model \cite{devlin-etal-2019-bert} to predict a triple plausibility. CoKE \cite{wang2019:coke} also tackles KG completion, as Masked Language Model (MLM) task, using a stack of Transformer blocks as encoder. To capture graph context, HittER \cite{chen-etal-2021-hitter} utilizes two levels of Transformer blocks, one to provide relation-dependent embeddings for an entity's neighbors and the other to aggregate their information. Despite their high-dimensional embeddings, these two methods, especially Hitter, are large and impose too many free parameters that cause considerable time and memory complexity.

Neural network methods share learned knowledge through free parameters of the model, which enables them to attain more expressive representations and
reduce the dimensionality of embeddings with less performance degradation. However, the results of these methods are most often notably worse in low-dimensional embeddings. Therefore, scalability is still a significant issue for them.

Out of the above main categories, some other methods such as MINERVA \citep{das2018go}, a reinforcement learning model, and DRUM \citep{NEURIPS2019_0c72cb7e}, a rule-based model, are proposed and reached interesting results. 

\section{Methodology}
\label{sec:methodology}
Despite the recent improvements on low dimensional KG embeddings, the considerable gap between the effectiveness of models in low dimensions and high dimensions has currently remained. This gap inspired us to exploit multi-head self-attention mechanism of Transformers as a general powerful method to effectively encode KGs in notably low dimensions and solve the link prediction problem. We aim to apply self-attention to our task since it is proven to be able to effectively capture dependencies and interactions within a sequence and produce highly expressive feature representations \citep{shen2019mutual, likhosherstov2021expressive}. In the following, we first introduce the formulation of KG completion, then elaborate on the proposed model, SAttLE.

\subsection{Background and Problem Formulation}
Let \(\pazocal{V}\) and \(\pazocal{R}\) represent the set of all entities and relations respectively. In addition, a triple is represented as \((s, r, t)\), where \(s, t \in \pazocal{V}\) are the source and target entities and \(r \in \pazocal{R}\) is the corresponding relation between them. A triple is true if it exists in a world or false either. A knowledge graph \(\pazocal{KG}\) is a subset of all possible true triples. Formally, each entity and relation in \(\pazocal{KG}\) is associated with an embedding vector.

\paragraph{Link prediction}In the context of KGs with relational data, the purpose of link prediction is to find true triples which are not observed in one \(\pazocal{KG}\) and add them to it. Concretely, given a triple \((s, r, t)\) that is not in the \(\pazocal{KG}\), the problem is to validate if it is true or not.

\subsection{SAttLE}

\begin{figure}[t]
\includegraphics[scale=0.25]{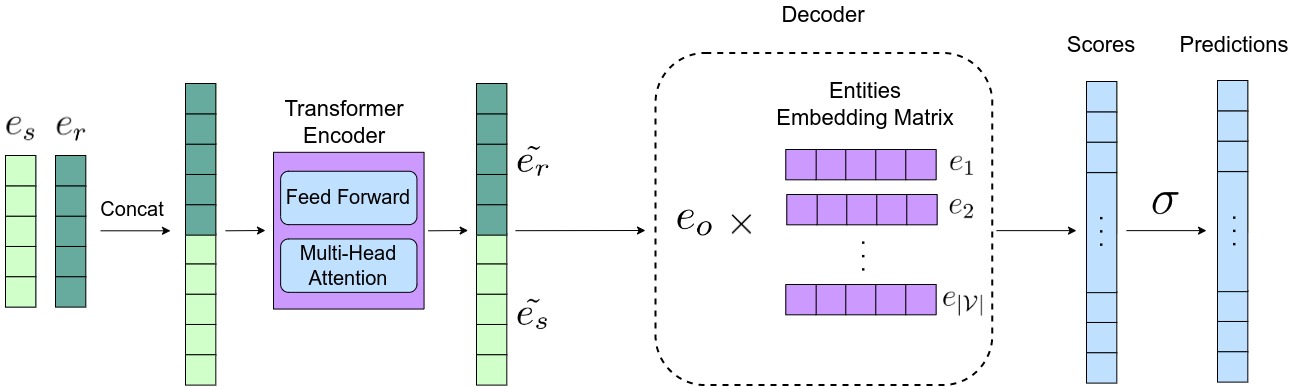}
\centering
\vspace{0.3cm}
\caption{
The architecture of SAttLE. (1) Source entity embedding $e_s$ and relation embedding $e_r$ are concatenated and fed to Transformer-based encoder, (2) representations $\tilde{e_s}$ and $\tilde{e_r}$ are generated respectively. (3) A decoder scores triples. First $e_o$ is created which in case of TwoMult decoder is equal to $\tilde{e_r}$ and in the case of Tucker decoder is the projection of $\tilde{e_s}$ made by the core tensor $W_c$, which its parameters are influenced by $\tilde{e_r}$. Then the scores are computed by multiplying $e_o$ and entities embedding matrix (4) Final probabilities are obtained by applying logistic sigmoid function to scores.
}
\label{fig:architecture}
\end{figure}

In summary, SAttLE takes the following steps: (1) assigns embedding vectors to each entity and relation in KG, and considers the pair \((s, r)\) of each triple \((s, r, t)\) as the input sequence and feeds it to a Transformer encoder, (2) generates highly expressive representations for \(r\) and \(s\) of the input sequence in low-dimensions, (3) scores the target of the corresponding triple \(t\) based on these representations, and (4) converts the score of $t$ to a probability by applying the logistic sigmoid to the score. A visualization of the steps in SAttLE is given in Figure \ref{fig:architecture}. Details of the above steps are explained in the following. 

The first step is assigning an embedding vector $e_v \in \mathbb{R}^{d_e}$ to each entity $v \in \pazocal{V}$, and an embedding vector $e_r \in \mathbb{R}^{d_r}$ to each relation \(r \in \pazocal{R}\). To preserve the compatibility of tensor calculus in our model, both relations and entities have a same dimensionality of embeddings (DoE), that is \(d_e=d_r=d\).

 For each triple \((s, r, t)\) where \(s\) has embedding \(e_s \in \mathbb{R}^{d}\) and \(r\) has embedding \(e_r \in \mathbb{R}^{d}\), the embedding matrix of the pair \((e_s, e_r)\) is considered to be the query \(Q \in \mathbb{R}^{2  \times d}\), the key \(K \in \mathbb{R}^{2 \times d}\), and the value \(V \in \mathbb{R}^{2  \times d}\) inputs to a transformer block \citet{NIPS2017_3f5ee243}. Assuming a simple one-head attention, the attention matrix \(A_{sr} \in \mathbb{R}^{2 \times 2}\) is computed as:
\begin{align}
\label{eq:attMatrix}
    A_{sr}=\text{softmax}(\frac{QK^T}{\sqrt{d_k}}).
\end{align}
Finally, the output representation matrix of the attention block \(H \in \mathbb{R}^{2 \times d}\) for the the source and relation is:
\begin{align}
\label{eq:reprsMatrix}
    H=A_{sr}V.
\end{align}
Attention matrix \(A_{sr}\) indicates how much should the source entity focus on its own information against the information of the relation and vice versa. As noted by \citet{likhosherstov2021expressive}, we can interpret \(A_{sr}\) as a query-dependent linear projection that applies to the input and propagates mutual information between input elements through itself. Thus, we found it an efficient way to fuse information of an entity and a relation. 

\paragraph{Multi-head attention}Considering the importance of the attention matrix \(A_{sr}\) in our model, multi-head attention is the key to gaining expressive representations. Intuitively, multiple linear projections apply to query, key, and value by different sets of parameters resulting in multiple queries, keys, and values. Self-attention mechanism is conducted over each set of query, key, and value, then final value vectors are concatenated and linearly transformed. Mathematically speaking:
\begin{align}
\label{eq:multiHead}
    \text{MultiH}(Q,K,V)=[\text{H}{}_1,\dots,\text{H}{}_h]W_O,
\end{align}
where
\begin{align*}
   \text{H}{}_i=\text{softmax}(\frac{Q_iK_i^T}{\sqrt{d_k}})V_i.
\end{align*}
In the above equation, \(Q_i=QW_i^Q\), \(K_i=KW_i^K\), and \(V_i=VW_i^V\) where \(W_i^Q \in \mathbb{R}^{d \times d_k}\), \(W_i^K \in \mathbb{R}^{d \times d_k}\), \(W_i^V \in \mathbb{R}^{d \times d_v}\), and \(W^O \in \mathbb{R}^{hd_v \times d}\) are projection matrices and \(h\) is the number of heads. 

Like usual Transformer model of \citet{NIPS2017_3f5ee243}, a token-wise feed-forward is applied to produce the final output representation matrix \(H_o\).
\begin{align}
\label{eq:MultiReprsMatrix}
    H_o = FFN(\text{MultiH}(Q,K,V)),
\end{align}
where
\begin{align*}
    FFN(x) &= max(0, xW_1 + b_1)W_2 + b_2,
\end{align*}
wherein \(W_1 \in \mathbb{R}^{d \times d_{h}}\), \(b_1 \in \mathbb{R}^{d_{h}}\), \(W_2 \in \mathbb{R}^{d_{h} \times d}\), and \(b_2 \in \mathbb{R}^{d}\) are trainable parameters. Unlike self-attention, the token-wise feed-forward transforms all elements in the sequence (two element in our case) independently without considering their interactions. Noteworthy, \citet{Yun2020Are} proves that the value-mapping ability of the token-wise feed-forward in combination with self-attention makes Transformers universal approximators.

\subsection{Decoding}
So far, we have just efficiently represented the source and the relation in a low-dimensional space. The next task is decoding which aims at incorporating the information embedded in the target entity to score the triple \((s, r, t)\). We are choosing two kinds of effective and simple decoding methods. It could be possible to get better results by using more complex decoding methods. Here, we select simple decoding methods and leave the others for interested researchers in future works. 

In the following, we present our approaches to score each triple \((s, r, t)\) for a possible link between. Consider  \(H_o=(\tilde{e}_s;\tilde{e}_r)\) is the encoder output representation, where \(\tilde{e}_s \in \mathbb{R}^{d}\) and \(\tilde{e}_r \in \mathbb{R}^{d}\) are the output representations of the \(s\) and \(r\), respectively.

\paragraph{TwoMult}
This simple method is inspired by the method of \citep{DBLP:journals/corr/YangYHGD14a}. Given a triple \((s, r, t)\), the score is computed by naively computing the two-way inner product between the representation of the relation \(\tilde{e}_r\) and the embedding of the target entity \(e_t\):
\begin{align}
\label{eq:TwoMult}
    \phi(s, r, t) = \tilde{e}_r^Te_t
\end{align}
For the rest of this paper, we refer to this method of scoring as TwoMult standing for Two-way Multiplication. Despite its simplicity, we believe that TwoMult can be an efficient way of decoding the embedded features. The logic behind it is that the model effectively embeds the information of the source entity into the relation representation in an implicit manner. As a result, the relation output representation would be descriptive enough to utilize solely. Our experimental results confirm this intuition.

Note that we also examined using the output representation of the source entity instead of the relation. Interestingly, the MRR metric, see definition in  \ref{subappendix:evalProtocol}, reduced by almost 1.5 percent by this replacement. Reasonably, we found that in our model the output representation of the relation carries more important information than the source entity.

\paragraph{Tucker}
Recently, the state-of-the-art tensor decomposition model TuckER \citep{balazevic-etal-2019-tucker} has gained promising results in link prediction on KGs. Considering a \(\pazocal{KG}\) represented as a binary tensor, TuckER uses Tucker decomposition \citep{tucker1966some} to compute a validation score for each triple as:
\begin{align}
\label{eq:TuckerOrg}
    \phi(s, r, t) = W_c\times_1e_s\times_2e_r\times_3e_t,
\end{align}
where \(W_c \in \mathbb{R}^{d_e \times d_r \times d_e}\) is the core tensor of the decomposition which can be learned and \(\times_n\) indicates n-mode tensor product. In SAttLE, we reuse the original TuckER in the following form with some changes in the model configuration (see  \ref{subappendix:modelSetups} for details):
\begin{align}
\label{eq:TuckerSelf}
    \phi(s, r, t) = W_c\times_1\tilde{e}_s\times_2\tilde{e}_r\times_3e_t,
\end{align}
and \(W_c \in \mathbb{R}^{d \times d \times d}\). It is worthwhile to state that, the number of core tensor free parameters in the original TuckER scales exponentially with the size of embeddings. Since our embedding vectors are low-dimensional, the number of free parameters of the core tensor is not restrictive. 

Eventually to validate a triple \((s, r, t)\), we compute the probability of \(t\) to be a target of the pair \((s, r)\) by applying a logistic sigmoid function to its score computed by one of the mentioned decoding approaches.

\subsection{Training}
At inference time given a triple \((s, r, t)\) we expect the model to score true target \(t\) high among other entities for query \((s, r, ?)\) and score true source s high among other entities for query \((?, r, t)\). In order to satisfy this expectation, we train our model using reciprocal learning \citep{lacroix2018canonical}. Thus for each triple \((s, r, t)\) we add its inverse \((t, r^{-1}, s)\) to the dataset. Also, we utilize 1-N scoring method \citep{dettmers2018convolutional} and score each pair \((s, r)\), containing the inverse triples as well, with all entities as target \(t\). We optimize with Binary Cross-Entropy (BCE) loss to enforce model score observed (valid) triples higher than others. Accordingly, for one sample triple \((s, r, t)\) we have:
\begin{align}
\label{eq:BCELoss}
    \pazocal{L}=-\frac{1}{|\pazocal{V}|}\sum_{t' \in \pazocal{V}} y_{t'}\log{p_{t'}}&+(1-y_{t'})\log{(1-p_{t'})},
\end{align}
where
\begin{align*}
p_{t'}=\sigma(\phi(s,r,t'))
    \nonumber\text{  and }y_{t'}= \left\{ \begin{array}{rcl}
                1 & \text{if } t'=t, \\
                0 & \text{otherwise}.
            \end{array}\right.
\end{align*}
In the above equation, \(\sigma(.)\) is the logistic sigmoid function and \(\pazocal{V}\) is the set of all entities. Also, we think that using an advanced alternative of BCE like OSM (One-Sided Margin) proposed by \citet{https://doi.org/10.48550/arxiv.2206.01002}, which can guarantee the width of the prediction margin, could improve our performance. This could be pursued in future researches.

\begin{table}[t!]
	\centering
	\caption{Datasets statistics. The number of triples in training, test, and validation sets are shown. Also, $|\pazocal{V}|$ and $|\pazocal{R}|$ are the total number of entities and relations respectively.}
	\small
    \begin{tabular}{L{0.8in}R{0.4in}R{0.4in}R{0.6in}R{0.6in}R{0.6in}}
		\toprule
		Datasets & $|\pazocal{V}|$ & $|\pazocal{R}|$ & Training & Validation & Test \\
		\midrule
		FB15k-237 & 14541  & 237 & 272115 & 17535 & 20466 \\
		WN18RR & 40943 & 11 & 86835 & 3034 & 3134 \\
		\bottomrule
	\end{tabular}
	\label{tab:datasetStatistics}
	\vskip -0.1in
\end{table}

Furthermore, we observed that adding L2 regularization weakens our model performance. However, we anticipate adding other types of regularization approaches such as N3 regularization \citep{lacroix2018canonical} would be a good practice that can make notable improvements. Here, we do not add any regularization and let it as an opportunity for future works.

\section{Experiments}
Here, we compare SAttLE with the recent state-of-the-art models on two most commonly used benchmark datasets for link prediction to verify its effectiveness. Moreover, we also study the influence of the main components of our model on its performance. Check  \ref{subappendix:modelSetups} for model setups and  \ref{subappendix:parameterSetups} for hyper-parameter setups. We make our implementation publicly available.

\paragraph{Datasets}
We evaluate SAttLE on FB15k-237 \citep{toutanova2015representing} and WN18RR \citep{dettmers2018convolutional} which are subsets of FB15k and WN18, both by \citet{NIPS2013_1cecc7a7}. These subsets do not have the information leakage problem of the training dataset to the test and validation, and therefore they are more challenging than the original datasets. Consequently, we do not experiment on FB15k and WN18 because of this leakage problem as noted by \citet{dettmers2018convolutional}. Statistics of the datasets are summarized in Table \ref{tab:datasetStatistics}.

\paragraph{Baseline}
We compared our method against some strong baselines and the current state-of-the-art models of different categories. Among neural network approaches in Euclidean space, we selected ConvE \citep{dettmers2018convolutional}, A2N \citep{bansal2019a2n}, QuatE \citep{NEURIPS2019_d961e9f2} and two Transformer-based models CoKE \citep{wang2019:coke} and HittER \citep{chen-etal-2021-hitter}. Also, we take TuckER \citep{balazevic-etal-2019-tucker} and HypER \citep{balavzevic2019hypernetwork} as tensor factorization models in the Euclidean space. RotatE \citep{sun2018rotate} and ComplEx \citep{trouillon2016complex} are chosen as methods based on the complex space representation. Finally among hyperbolic space approaches, MuRP \citep{NEURIPS2019_f8b932c7} and ATTH \citep{chami-etal-2020-low} are considered.

\paragraph{Evaluation Protocol}
We evaluate our model for link prediction using two standard metrics which are commonly used: Mean Reciprocal Rank (MRR), and Hits@\(k\) ratios both with \textit{filtered} setting \citep{NIPS2013_1cecc7a7}. More information is provided in  \ref{subappendix:evalProtocol}.

\section{Results}
In order to evaluate SAttLE performance, we challenged it against some powerful state-of-the-art models. The empirical results on FB15k-237 and WN18RR are summarized in Table \ref{tab:comparisons}. The Dimensionality of Embeddings (DoE) for each model is also reported alongside its name. For ComplEx, Rotate, and QuatE, reported DoE is the sum of all assigned embedding vectors. For the case of ComplEx and Rotate, both real and complex components are counted for DoE. Similarly for the case of QuatE, real and all quaternions are counted. The QuatE results are reported without N3 regularization (check Appendix 7.2 of \citet{NEURIPS2019_d961e9f2}).

Table \ref{tab:comparisons} clearly brightens the great performance of SAttLE against the competitors with significantly lower dimensionality of embeddings. Overall, SAttLE has the second best performance on WN18RR. In addition, SAttLE beats most of the state-of-the-art methods on FB15k-237 except for CoKE and HittER.

\begin{table*}[t!]
	\caption{Link prediction results on FB15k-237 and WN18RR. Results of $\dagger$ are taken from \citep{sun2018rotate}. Other results are taken from original papers and their own experimental setups. DoE stands for Dimensionality of Embeddings. The best results are in bold and the second bests are underlined.}
	\vskip 0.1in
	\footnotesize
	\centerline{
    \begin{tabular}{L{1.7in}C{0.2in}*{8}{C{0.3in}}}
		\toprule
		\multirow{2}{4em}{} & \multirow{2}{4em}{\textbf{DoE}} & \multicolumn{4}{c}{\textbf{FB15k-237}} &
		\multicolumn{4}{c}{\textbf{WN18RR}} \\
		\cmidrule(lr){3-6} \cmidrule(lr){7-10}
		\rowfont{\scriptsize}%
		&  &   MRR   &   Hits@1   &   Hits@3   &   Hits@10   &   MRR   &   Hits@1   &   Hits@3   &   Hits@10 \\
		\midrule
		\rowfont{\footnotesize}%
		ComplEx$\dagger$ \citep{trouillon2016complex}  &    400   &   .247   &   .158   &   .275   &   .428   &   .44   &   .41   &   .46   &   .51 \\
		ConvE \citep{dettmers2018convolutional}   &   200   &   .325   &   .237   &   .356   &   .501   &   .43   &   .40   &   .44   &   .52 \\
		A2N \citep{bansal2019a2n}  &    512   &   .317   &   .232   &   .348   &   .486   &   .45   &   .42   &   .46   &   .51 \\
		MURP \citep{NEURIPS2019_f8b932c7}   &   200   &   .335   &   .243   &   .367   &   .518   &   .481   &   .44   &   .495   &   .566 \\
		RotatE \citep{sun2018rotate}   &   1000   &   .338   &   .241   &   .375   &   .533   &   .476   &   .428   &   .492   &   .571 \\
		HypER \citep{balavzevic2019hypernetwork}   &   200   &   .341   &   .252   &   .376   &   .52   &   .465   &   .436   &   .477   &   .522 \\
		QuatE \citep{NEURIPS2019_d961e9f2}   &   400   &   .348   &   .248   &   .382   &   \underline{.55}   &   .488   &   .438   &   \underline{.508}   &   \underline{.582} \\
		ATTH \citep{chami-etal-2020-low}   &   500   &   .348   &   .252   &   .384   &   .54   &   .486   &   .443   &   .499   &   .573 \\
		TuckER \citep{balazevic-etal-2019-tucker}   &   200   &   .358   &   .266   &   .394   &   .544   &   .47   &   .444   &   .482   &   .526 \\
		CoKE \citep{wang2019:coke}   &   256   &   \underline{.364}   &   \underline{.272}   &   \underline{.400}   &   .549   &   .484   &   .450   &   .496   &   .553 \\
		HittER \cite{chen-etal-2021-hitter}   &   320   &   \textbf{.373}   &   \textbf{.279}   &   \textbf{.409}   &   \textbf{.558}   &   \textbf{.503}   &   \textbf{.462}   &   \textbf{.516}   &   \textbf{.584} \\
		\midrule
		SAttLE-TwoMult   &   \underline{100}   &   .360   &   .268   &   .396   &   .545   &   \underline{.491}   &   \underline{.454}   &   \underline{.508}   &   .558 \\
		SAttLE-Tucker   &   \textbf{64}   &   .358   &  .266   &   .394   &   .541   &   .476   &   .442   &   .490   &   .54 \\
		\bottomrule
	\end{tabular}}
	\label{tab:comparisons}
\end{table*}

Obviously, on both datasets HittER has achieved the best performance. Actually, HittER is a powerful model composed of two transformer blocks to produce relation-dependent information for neighbors and capture graph structure. However, these two blocks have many parameters and increase time and memory complexity considerably. On the other hand, HittER and CoKE both use multiple encoder blocks same as \citep{NIPS2017_3f5ee243} which makes their complexity issue severe. 

Clearly, on FB15k-237, TuckER is our closest competitor. Multi-task learning by sharing knowledge across relations makes TuckER effective on datasets with relatively more relations. However on datasets like WN18RR with more entities and small number of relations, TuckER has relatively low performance. In contrast, self-attention in SAttLE captures dependencies and interactions between both entities and relations. Therefore, it performs well on both datasets with lots of entities or lots of relations. On the other hand, number of free parameters in SAttLE-Twomult grows linearly with respect to the dimension of the embeddings while it grows exponentially for TuckER because of its core tensor.

QuatE has the closets results to ours on WN18RR, since it efficiently models such datasets with numerous symmetry relations. On the other hand, WN18RR has lots of hierarchical relations that explains the good results of ATTH and MURP on this dataset because of using embeddings in the hyperbolic domain. To the best of our knowledge these two models are the only models that concern low-dimensional embeddings to tackle scalability issues so we made comparisons between them and SAttLE with embeddings of very low dimensions.

Table \ref{tab:lowComparisons} shows performance of ATTH, MURP, and SAttLE in very low dimensions ($d=32$). As discussed above, performances of MURP and ATTH oscillate between different datasets depending on the number of hierarchical relations. Therefore, they perform worse on FB15k-237 with fewer hierarchical relations compared to WN18RR although they have better results on WN18RR. It also worth mentioning that despite their promising results in low-dimensions, their best results are achieved in high-dimensions and SAttLE surpass them in higher dimensions as shown in Table \ref{tab:comparisons}. Besides, SAttLE makes no assumptions about the data making it efficiently applicable on different datasets.

\begin{table*}[t!]
	\caption{Results on FB15k-237 and WN18RR with $d=32$. Both MURP and ATTH results are taken from \citep{chami-etal-2020-low}. The best results are in bold.}
	\vskip 0.1in
	\footnotesize
	\centerline{
    \begin{tabular}{L{1in}*{8}{C{0.3in}}}
		\toprule
		\multirow{2}{4em}{} & \multicolumn{4}{c}{\textbf{FB15k-237}} &
		\multicolumn{4}{c}{\textbf{WN18RR}} \\
		\cmidrule(lr){2-5} \cmidrule(lr){6-9}
		\rowfont{\scriptsize}%
		&  MRR   &   Hits@1   &   Hits@3   &   Hits@10   &   MRR   &   Hits@1   &   Hits@3   &   Hits@10 \\
		\midrule
		\rowfont{\footnotesize}%
		MURP  &   .323   &   .235   &   .353   &   .501   &   .465   &   \textbf{.420}   &   \textbf{.484}   &   .544 \\
		ATTH  &   .324   &   .236   &   .354   &   .501   &   \textbf{.466}   &   .419   &   \textbf{.484}   &   \textbf{.551} \\
		SAttLE-Tucker  &    \textbf{.340}   &   \textbf{.252}   &   \textbf{.372}   &   \textbf{.513}   &   .454   &   .414   &   .474   &   .527\\
		\bottomrule
	\end{tabular}}
	\label{tab:lowComparisons}
\end{table*}

Lastly, The main goal of this work is to decrease the dimensionality of embeddings while retaining the model effectiveness. The empirical results in Table \ref{tab:comparisons} clearly demonstrate how notably efficient SAttLE is with embeddings in comparably lower dimensions than other methods.

Considering both datasets, SAttLE outperforms or performs competitively with HittER, CoKE, QuatE, TuckER, and ATTH as the five best state-of-the-arts which have respectively 3.2x, 2.56x, 4x, 2x, and 5x, or in average 3.35x, greater number of dimensions. Consequently, SAttLE is highly expressive and successfully alleviates the scalability problem for huge KGs with a vast number of entities and relations.

\subsection{Ablation Study}
\subsubsection{Effect of the number of attention-heads}
\label{sec:atTheads}
Using self-attention to obtain powerful representations by absorbing the information interaction and dependencies between the entities and relations is the core idea of our model. We discussed that multi-head attention is beneficial to produce representations in different subspaces and encode different dependencies and relationships by attending to each representation space. 

Previously HittER and CoKE have used multiple encoder blocks with low number of attention heads same as the original Transformer. However, we found that a sufficiently large number of heads results in more expressiveness than using multiple encoder blocks. More detailed comparison in Section \ref{sec:paramEff}.

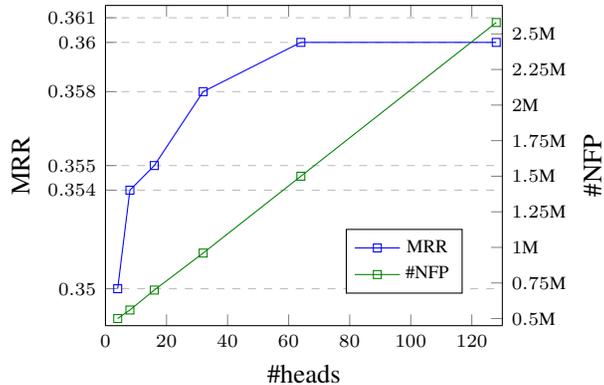
\begin{figure*}[t!]
    \centering
    \vskip 0.1in
    \begin{tikzpicture}
    \begin{axis}[
        width=2.7in, height=2.3in,
        xlabel={\#heads},
        xlabel near ticks,
        ylabel={MRR},
        ylabel near ticks,
        xmin=0, xmax=130,
        ymin=0.3485, ymax=0.3615,
        axis y line*=left,
        xtick={0,20,40,60,80,100,120},
        ytick={0.35, 0.354, 0.355, 0.358, 0.36, 0.361},
        xticklabel style={font=\scriptsize},
        yticklabel style={
            font=\scriptsize,
            scaled ticks=false,
            /pgf/number format/fixed,
            /pgf/number format/precision=3},
        ymajorgrids=true,
        grid style=dashed,
    ]
    \addplot[color=blue, mark=square, mark options={scale=0.75}]
        coordinates {
        (4, 0.35)
        (8, 0.354)
        (16, 0.355)
        (32, 0.358)
        (64, 0.36)
        (128, 0.36)
        };
    \label{MRR}
    \end{axis}
    \begin{axis}[
        width=2.7in, height=2.3in,
        ylabel={\#NFP},
        ylabel near ticks,
        xmin=0, xmax=130,
        ymin=0.45, ymax=2.7,
        hide x axis,
        axis y line*=right,
        ytick={0.5,0.75,1,1.25,1.5,1.75,2,2.25,2.5},
        yticklabel style = {font=\scriptsize},
        yticklabel=$\pgfmathprintnumber{\tick}$M,
        grid style=dashed,
        legend style={font=\scriptsize},
        legend style={
            at={(0.9,0.1)},
            anchor=south east,
            font=\scriptsize
        }
    ]
    \addlegendimage{/pgfplots/refstyle=MRR}\addlegendentry{MRR}
    \addplot[color=darkGreen, mark=square, mark options={scale=0.75}]
        coordinates {
        (4, 0.5)
        (8, 0.56)
        (16, 0.70)
        (32, 0.96)
        (64, 1.50)
        (128, 2.58)
        };
    \addlegendentry{\#NFP}
    \end{axis}
    \end{tikzpicture}
    \caption{MRR of SAttLE-TwoMult on FB15k-237 and the Number of Nonembedding Free Parameters (NFP) of the model for different number of attention heads for $d=100$ and \#heads \(\in \{4, 8, 16, 32, 64, 128\}\). Sufficiently large \#heads makes improvements without adding lots of parameters. Even with low \#heads and negligible free parameters SAttLE is so effective.}
    \label{fig:attentionHeads}
    \vskip -0.1in
\end{figure*}

The outcomes in Figure \ref{fig:attentionHeads} clearly confirm our hypothesis that enhancing the number of heads offers better representations and makes advantages without overfitting. Meanwhile, it illustrates linear growth of the number of Nonembedding Free Parameters (NFP) by increasing the number of attention heads. In consequence, we make a trade-off and choose 64 heads since it leads to desirable performance while avoids enlarging the model. As declared in Figure \ref{fig:attentionHeads}, we find that self-attention with small number of heads and negligible number of free parameters can also offer great advantages.

\begin{table}[b!]
	\caption{Results on FB15k-237 using different DoEs for our two decoders, namely TwoMult and Tucker.}
	\vskip 0.1in
	\footnotesize
	\centerline{
    \begin{tabular}{L{1.1in}C{0.4in}*{4}{C{0.3in}}}
		\toprule
		\rowfont{\scriptsize}%
	                         &  DoE ($d$) &   MRR  & Hits@1 & Hits@3 & Hits@10 \\
		\midrule
		\rowfont{\footnotesize}%
		\multirow{3}{8em}{SAttLE-TwoMult}  &  100  &  .360  &  .268  &  .396  & .545 \\
                                    &  64   &  .355  &  .264  &  .389  & .535 \\
                                    &  32   &  .335  &  .247  &  .367  & .507 \\
        \midrule
		\multirow{3}{8em}{SAttLE-Tucker}   &  100  &  .357  &  .265  &  .395  & .541 \\
                                    &  64   &  .358  &  .266  &  .393  & .541 \\
                                    &  32   &  .340  &  .252  &  .372  & .513 \\
		\bottomrule
	\end{tabular}}
	\label{tab:embeddings}
\end{table}

\subsubsection{Effect of the dimensionality of embeddings}
We have also challenged SAttLE in very low dimensions to demonstrate its great scalability to very big KGs. Table \ref{tab:embeddings} shows SAttLE performance in different dimensionality of embeddings. Notably in \(d=64\), SAttLE surpasses most of our competitors. The model efficiency is also considerable in \(d=32\). Consequently, SAttLE can be an applicable and expressive choice in the cases where numerous entities or relations make memory and computation bottlenecks.

\subsubsection{Effect of the decoding methods}
SAttLE implicitly encodes the information of the source entity in the relation before decoding. In consequence, our experiments in Table \ref{tab:comparisons} demonstrate that the simple TwoMult method interestingly performs quite well and reduces the risk of overfitting. Especially, TwoMult scoring method is powerful and useful when a considerably low memory rules decision making on model architecture since it adds no more parameters.

As discussed before, SAttLE is easily adaptive with different decoding methods. Table \ref{tab:embeddings} shows that applying a more complex decoding method like Tucker would make considerable improvements in very low dimensions, for example  \(d=32\). 

However, SAttLE performance with Tucker decoder degrades in \(d=100\). We think a suitable regularization method or just a wider hyper-parameter search could refine this behavior. In contrast, TwoMult simplicity brings the advantage of better generalization in \(d=100\) to attain great results.

\begin{table}[t!]
	\caption{Parameter efficiency on FB15k-237 and WN18RR.}
	\vskip 0.1in
	\footnotesize
	\centerline{
    \begin{tabular}{L{1.3in}*{4}{C{0.4in}}}
		\toprule
		\multirow{2}{4em}{} & \multicolumn{2}{c}{\textbf{FB15k-237}} &
		\multicolumn{2}{c}{\textbf{WN18RR}} \\
		\cmidrule(lr){2-3} \cmidrule(lr){4-5}
		\rowfont{\scriptsize}%
		&  \#NFP   &   \#EFP    &   \#NFP   &   \#EFP  \\
		\midrule
		\rowfont{\footnotesize}
		CoKE   &  6.16M   &  3.84M   &   6.51M   &   10.49M \\
		HittER   &   11.20M   &  4.80M    &   10.89M   &  13.11M  \\
		\midrule
		SAttLE-TwoMult   &   \textbf{1.50M}   &  \textbf{1.50M}   &   \textbf{1.14M}   &   \textbf{4.10M}  \\
		\bottomrule
	\end{tabular}}
	\label{tab:paramEff}
\end{table}

\subsection{Discussion}
\subsubsection{Parameter efficiency of SAttLE}
\label{sec:paramEff}
As discussed before, compared to previous Transformer-based models SAttLE utilizes single encoder block with large number of attention heads rather than multiple encoders blocks. Table \ref{tab:paramEff} illustrates the parameter efficiency of SAttLE based on the number of Nonembedding Free Parameters (NFP) and Embedding Free Parameters (EFP). 

Obviously, our model is remarkably smaller than CoKE and HittER with 75\% and 87\% lower \#NFP on FB15k-237, also 82\% and 89\% \#NFP on WN18RR respectively. This confirms our claim that instead of multiple encoder blocks and low number of heads, as used in CoKE and HittER, we can decrease model complexity and gain competitive performance by one encoder block and increasing the number of heads. 

On the other hand, as discussed before, we have considerably lower \#EFP because of lower dimensionality of embeddings. Especially, these two parameter reductions are more important and significant in the case of immense KGS with numerous entities and relations.

\subsubsection{Time Complexity}
As we know, SAttLE is consist of a Transformer-based encoder and two different decoding methods. The multi-head attention block in our model takes three steps. First, to produce matrices 
$Q \in \mathbb{R}^{2\times d}$, $K \in \mathbb{R}^{2\times d}$
, and 
$V \in \mathbb{R}^{2 \times d}$ 
in each head, input matrix 
$(e_s; e_r) \in \mathbb{R}^{2 \times d}$ 
should be linearly transformed by three learned matrices in 
$\mathbb{R}^{d \times d}$ 
which has computational complexity $O(d^2)$. Second to compute the output representation matrix 
$H \in \mathbb{R}^{2\times d}$ 
from Equation \ref{eq:reprsMatrix}, we have to compute the attention matrix 
$A_{sr} \in \mathbb{R}^{2 \times 2}$ 
from Equation \ref{eq:attMatrix} and that takes the time complexity of $O(d)$ overall. Third step is to compute 
$H_o \in \mathbb{R}^{2d}$ 
from Equation \ref{eq:MultiReprsMatrix} which also has complexity $O(d)$. Therefore, time complexity of our encoder is $O(d^2)$. Regrading Equation \ref{eq:TwoMult}, if we use TwoMult then the decoder complexity is $O(d)$ because we just have a simple multiplication. On the other hand, using Tucker as decoder imposes complexity $O(d^3)$ considering Equation \ref{eq:TuckerSelf}. Consequently, SAttLE-TwoMult and SAttLE-Tucker have time complexity $O(d^2)$ and $O(d^3)$, respectively.

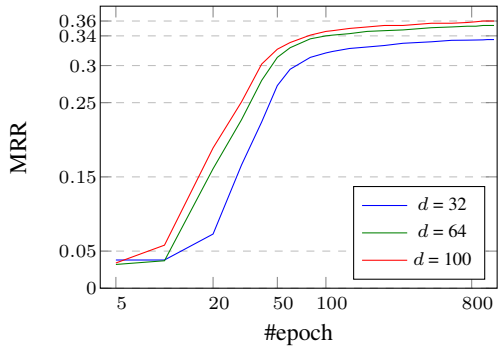
\begin{figure}[b!]
    \centering
    \begin{tikzpicture}
        \begin{axis}[
            width=2.7in, height=2.1in,
            xlabel={\#epoch},
            ylabel={MRR},
            xlabel style={
                yshift = {5}, font=\small
            },
            ylabel style={
                yshift = {-4}, font=\small
            },
            xmin=4, xmax=1150,
            ymin=0.0, ymax=0.38,
            xtick={2,5,20,50,100,800},
            ytick={0.0,0.05,0.15,0.25,0.30,0.34,0.36},
            /pgf/number format/1000 sep={},
            xmode = log,
            log basis x={4},
            xticklabel={%
            \pgfmathparse{int(4^\tick)}%
            \pgfmathprintnumber{\pgfmathresult}
            },
            yticklabel style={
            scaled ticks=false,
            /pgf/number format/fixed,
            /pgf/number format/precision=3},
            every tick label/.append style={font=\scriptsize, xshift=0.5ex},
            clip=false,
            legend pos=south east,
            legend style={font=\scriptsize},
            ymajorgrids=true,
            grid style=dashed,
        ]
        \addplot[color=blue]
            coordinates {
            (5, 0.038)
            (10, 0.038)
            (20, 0.073)
            (30, 0.166)
            (40, 0.224)
            (50, 0.273)
            (60, 0.295)
            (80, 0.311)
            (100, 0.317)
            (110, 0.319)
            (140, 0.323)
            (180, 0.325)
            (230, 0.327)
            (300, 0.33)
            (450, 0.332)
            (600, 0.334)
            (750, 0.3342)
            (850, 0.3344)
            (940, 0.3346)
            (1000, 0.335)
            (1100, 0.335)
            };
            \addlegendentry{$d$ = 32}
        \addplot[color=darkGreen]
            coordinates {
            (5, 0.032)
            (10, 0.037)
            (20, 0.161)
            (30, 0.227)
            (40, 0.280)
            (50, 0.311)
            (60, 0.324)
            (80, 0.336)
            (100, 0.340)
            (110, 0.341)
            (140, 0.343)
            (180, 0.346)
            (230, 0.347)
            (300, 0.348)
            (450, 0.351)
            (600, 0.352)
            (750, 0.353)
            (850, 0.353)
            (940, 0.354)
            (1000, 0.354)
            (1100, 0.354)
		    };
            \addlegendentry{$d$ = 64}
        \addplot[color=red]
            coordinates {
            (5, 0.034)
            (10, 0.058)
            (20, 0.189)
            (30, 0.251)
            (40, 0.302)
            (50, 0.322)
            (60, 0.331)
            (80, 0.341)
            (100, 0.346)
            (110, 0.347)
            (140, 0.35)
            (180, 0.352)
            (230, 0.354)
            (300, 0.354)
            (450, 0.357)
            (600, 0.357)
            (750, 0.358)
            (850, 0.359)
            (940, 0.36)
            (1000, 0.36)
            (1100, 0.36)
	        };
            \addlegendentry{$d$ = 100}
        \end{axis}
        \label{fig:MRRvsEpochs}
    \end{tikzpicture}%
    \caption{Convergence behavior of SAttLE-TwoMult on FB15k-237 for different DoEs $d$ \(\in \{32, 64, 100\}\).}%
    \label{fig:convergence}
\end{figure}

\subsubsection{Convergence Behavior}
KGs are intrinsically very dynamic, as a result in lots of real world applications multiple models should be trained based on the amount of data changes in a period of time. So, the convergence speed of models developed for link prediction is important. Figure \ref{fig:convergence} shows the convergence behavior of SAttLE for different DoEs on FB15k-237. Obviously, with small number of epochs, SAttLE achieves sufficiently desirable performance.

\section{Conclusion and Future Works}
We introduce SAttLE, a model powered by self-attention to embed KGs in low dimensions. Our Transformer-based method, efficiently models dependencies and interactions between entities and relations using multi-head self-attention mechanism to produce highly expressive representations. We utilized simple scoring methods to decode the representations. Regarding the empirical results on two standard link prediction datasets, SAttLE requires embeddings in significantly lower dimensions than the recent state-of-the-art models to achieve favorably comparable results on FB15k-237 and WN18RR.

There would be several extensions of our work that future researchers can look into them. The first is to take advantage of our encoder expressive representations and attach more efficient decoding methods. We did not use any regularization methods, so adding an appropriate regularization method such as the one proposed by \citet{lacroix2018canonical} could lead to considerable improvements. Lastly, we believe that one could just inclusively fine-tune the vast number of our model hyper-parameters, to explore a proper set and get better results.

\bibliographystyle{unsrtnat}
\bibliography{references}

\appendix
\section{Experimental Details}
\label{appendix:experimentalSetups}
\subsection{Model Setups}
\label{subappendix:modelSetups}
Following the settings of \citet{NIPS2017_3f5ee243} for the original Transformer model, we use dropout before and after the multi-head attention block, after applying softmax in each self-attention block, and also after the position feed-forward block with rates of do${}_{enc}$, do${}_{mha}$, do${}_{sdp}$, and do${}_{pff}$ respectively.

Moreover, compared to the original setting we remove the first fully-connected layer before the multi-head attention block since we observed that it slightly degrades the model performance. We conjecture that in our case, this projection layer might cause a non-optimal information fusion between the input relation and entity due to the difference between their embedding spaces. By the same intuition, the layer normalization before the multi-head attention block is respectively replaced by two separate batch normalizations, one for relations and one for entities. 

As the last configuration, we remove the layer normalization after the position-wise feed-forward when Tucker is used as the scoring function. Also, compared to the setting of \citet{balazevic-etal-2019-tucker} for TuckER, we preserve the first batch normalization on the source entity and remove all other batch normalizations and dropouts. 

\subsection{Hyper-parameter Setups}
\label{subappendix:parameterSetups}
We implemented our model in PyTorch \citep{paszke2019pytorch} and trained it on a single GPU. To optimize our model we use Adam \citep{DBLP:journals/corr/KingmaB14}. It is noteworthy to state that inheriting the characteristics of the original Transformers, our model has lots of hyper-parameters. Consequently, we could not search for all hyper-parameters at once. Instead, we fixed the number of attention heads to 64 as discussed in Section \ref{sec:atTheads} and searched for other hyper-parameters based on the validation sets performance according to MRR in the following manner.

We find all the hyper-parameters of the model with random search. First we tuned training parameters and fixed them for the rest of tunings. We chose batch size from \{128, 512, 1024, 2048\}, learning rate from \{0.0001, 0.0005, 0.001\}, learning rate decay from {1, 0.995} with fixed step size of 2 epochs. Second, we tuned mode architecture hyper-parameters except number of attention heads where $d_k$ and $d_v$ varied between \{32, 50, 64\} and $d_h$ between \{100, 512, 1024, 2048\}. 

We fixed previous hyper-parameters and tuned dropout rates in \{0.1, 0.2, 0.3, 0.4\} using Bayesian optimization combined with ASHA \citep{50611}, an asynchronous aggressive early stopping method. Table \ref{tab:settingPrameters} lists the full set hyper-parameters we chose to get the reported results of this paper. 

\subsection{Evaluation Protocol}
\label{subappendix:evalProtocol}
To evaluate link prediction in KGs, we use two standard metrics which are commonly used: Mean Reciprocal Rank (MRR), and Hits@\(k\) ratios. Both of these metrics are basically based on the ranking of the ground truth triple score against all of the corrupted triples. Specifically, we use \textit{filtered} setting \citep{NIPS2013_1cecc7a7}, so to compute left rank, we first perturb target entity of a triple with all entities that don’t make a corrupted triple which already exists in train, test, or validation sets except the true triple itself. Then, all corrupted triples are scored and sorted in descending order and the rank of the true triple is saved. We follow the same procedure to compute right rank of each triple by perturbing source entity. Take note that, for our results to be fair for comparison we use the random protocol proposed by \citet{sun-etal-2020-evaluation}, as a result, the true triple is randomly placed among corrupted triples before scoring and sorting. MRR is the average of the inverse of the mean ranks (over left and right ranks) of the true triples. Hits@\(k\) ratios are the percentage of times a true triple is ranked lower than or equal to $k$.

\begin{table*}[t]
	\caption{Hyper-parameters settings for the reported results in this paper.}
	\vskip 0.1in
	\centering
	\footnotesize
    \begin{tabular}{L{0.5in}L{0.4in}*{9}{R{0.2in}}*{4}{R{0.1in}}}
		\toprule
		           Datasets          &          Decoder           & $d$ & do${}_{enc}$  & do${}_{mha}$ & do${}_{pff}$ & do${}_{sdp}$ & $d_{h}$ & bs & dr & lr & ls & $d_k$ & $d_v$ & \#heads\\
		\midrule
		\multirow{6}{*}{FB15k-237} & \multirow{3}{*}{TwoMult} & 100 & 0.4   & 0.3  & 0.2  & 0.1  & \multirow{5}{*}{2048} & \multirow{5}{*}{2048} & \multirow{5}{*}{0.995} & \multirow{7}{*}{0.001} & \multirow{7}{*}{0.1} & \multirow{7}{*}{32} & \multirow{7}{*}{50} & \multirow{7}{*}{64}\\
		                           &                          & 64  & 0.2 & 0.3  & 0.2  & 0.1  & & & & & & & &\\
		                           &                          & 32  & 0.1 & 0.1  & 0.1  & 0.1  & & & & & & & &\\
		\cmidrule(lr){2-7}
		                             & \multirow{2}{4em}{Tucker}  & 100  & 0.4 & 0.2  & 0.3  & 0.1  & & & & & & & &\\
		                             &                            & 64  & 0.4 & 0.2  & 0.2  & 0.1  & & & & & & & &\\
		                             &                            & 32  & 0.1 & 0.2 & 0.1 & 0.1  & & & & & & & &\\
        \cmidrule(lr){1-10}
		\multirow{3}{*}{WN18RR}    & TwoMult                    & 100 & 0.3 & 0.4 & 0.4 & 0.1 & \multirow{3}{*}{100} & \multirow{3}{*}{1024} & \multirow{3}{*}{1} & & & & & \\
		                           & Tucker                     & 64  & 0.3 & 0.4 & 0.4 & 0.1 & & & & & & & & \\
		                           & Tucker                     & 32  & 0.1 & 0.1 & 0.3 & 0.4 & & & & & & & & \\
		\bottomrule
	\end{tabular}
	\label{tab:settingPrameters}
	\vskip -0.1in
\end{table*}

\end{document}